\documentclass[fleqn,11pt,bibbackend=biber,oneside]{FBItum}

\usepackage{tikz}
\newcommand{\ExternalLink}{%
    \tikz[x=1.2ex, y=1.2ex, baseline=-0.05ex]{%
        \begin{scope}[x=1ex, y=1ex]
            \clip (-0.1,-0.1) 
                --++ (-0, 1.2) 
                --++ (0.6, 0) 
                --++ (0, -0.6) 
                --++ (0.6, 0) 
                --++ (0, -1);
            \path[draw, 
                line width = 0.5, 
                rounded corners=0.5] 
                (0,0) rectangle (1,1);
        \end{scope}
        \path[draw, line width = 0.5] (0.5, 0.5) 
            -- (1, 1);
        \path[draw, line width = 0.5] (0.6, 1) 
            -- (1, 1) -- (1, 0.6);
        }
    }

\begin{document}
\selectlanguage{english} 
\parskip 8pt
\pagestyle{fbiplain}
\pagenumbering{arabic} %


\pptitle{Investigating Robot Dogs for Construction Monitoring: A Comparative Analysis of Specifications and On-site Requirements}

\ppauthor[1]{Miguel A.} {Vega Torres*}{\email{miguel.vega@tum.de}}
\ppauthor[1]{Fabian} {Pfitzner*}{\email{fabian.pfitzner@tum.de}}

\begin{affiliations}
  \ppaffil[1]{Chair of Computational Modeling and Simulation, TU~Munich, Arcisstr.~21, 80333~Munich, Germany}
\end{affiliations}

\begin{abstract}[BIM, Robotics, construction site, indoor monitoring, LiDAR, automation]

Robot dogs are receiving increasing attention in various fields of research. However, the number of studies investigating their potential usability on construction sites is scarce.  

The construction industry implies several human resource-demanding tasks such as safety monitoring, material transportation, and site inspections. 
Robot dogs can address some of these challenges by providing automated support and lowering manual effort.  

In this paper, we investigate the potential usability of currently available robot dogs on construction sites in terms of focusing on their different specifications and on-site requirements to support data acquisition.  
In addition, we conducted a real-world experiment on a large-scale construction site using a quadruped robot.  

In conclusion, we consider robot dogs to be a valuable asset for monitoring intricate construction environments in the future, particularly as their limitations are mitigated through technical advancements.
\end{abstract}
\ppcopyrightanddoi{will/be/assigned/by/editor}

\section{Introduction}
Research considering improving digitization on construction sites has increased significantly within the last years. 
An ongoing challenge is to acquire periodic data of the entire construction site for creating a comprehensive digital twin.

Robot dogs are designed based on the structure and motion of quadruped animals, featuring four mechanical legs for versatile terrain use.
Construction managers face time-consuming tasks to ensure project progress, often requiring manual site inspections. 
Autonomous robot-based construction monitoring offers a promising solution to reduce effort. 
While most research focuses on unmanned aerial vehicles (UAVs) for construction, ground-based legged robots could be better suited for the job \cite{halder2023robots}. 
To explore future research possibilities, the potential usability of robot dogs on construction sites needs investigation, considering their data acquisition capabilities, access to hard-to-reach areas, and ability to sustain with limited power supply options.

Overall, this paper offers (a) a detailed comparison of legged robots; (b) presents a compact mapping system for unmanned ground vehicles (UGVs); (c) discusses the potential for autonomous digital twin creation in construction, and identifies the challenges that need to be addressed for successful implementation. These contributions support advancing the use of robotics and mapping technologies in construction monitoring and management.

\section{Background}
The construction industry differs from controlled industrial environments, and its objective is to control on-site factors for economically optimal building production. 
Construction sites require diverse and dynamic monitoring approaches due to their unique nature. 
The application of robots in construction has increased, with companies using them to solve labor-intensive and hazardous tasks. \cite{halder2023robots}.
Legged robots are preferred for navigating challenging terrains found on construction sites. 
Some studies have explored using legged robots for progress monitoring and safety management, particularly in tracking scaffolding \cite{kim2022deep}.
The suitability and practicality of legged robot systems on construction sites remain open topics. 
This paper analyzes available quadruped-legged robots for construction monitoring and presents a case study of automatic data acquisition using a self-developed mapping system.


\section{Quadruped robots}

In this section, we demonstrate a comprehensive list of legged robots available in the European market together with their main characteristics and current prices. 
Due to relevance and usability, we do not include robots heavier than 50 Kg or smaller than 40 cm tall.

\begin{figure}%
    \centering
    \subfloat[A1]{{\includegraphics[width=2.5cm]{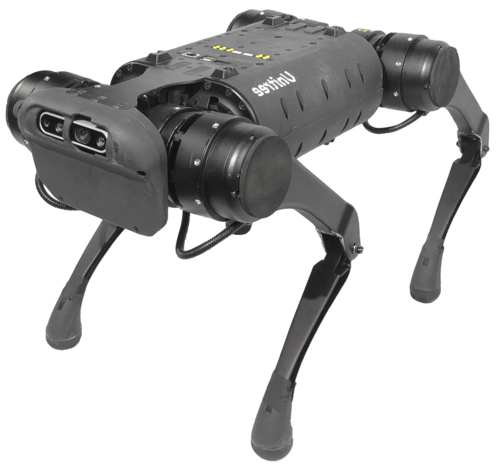}} 
    }\label{fig:A1}
    \subfloat[Go1]{{\includegraphics[width=2.5cm]{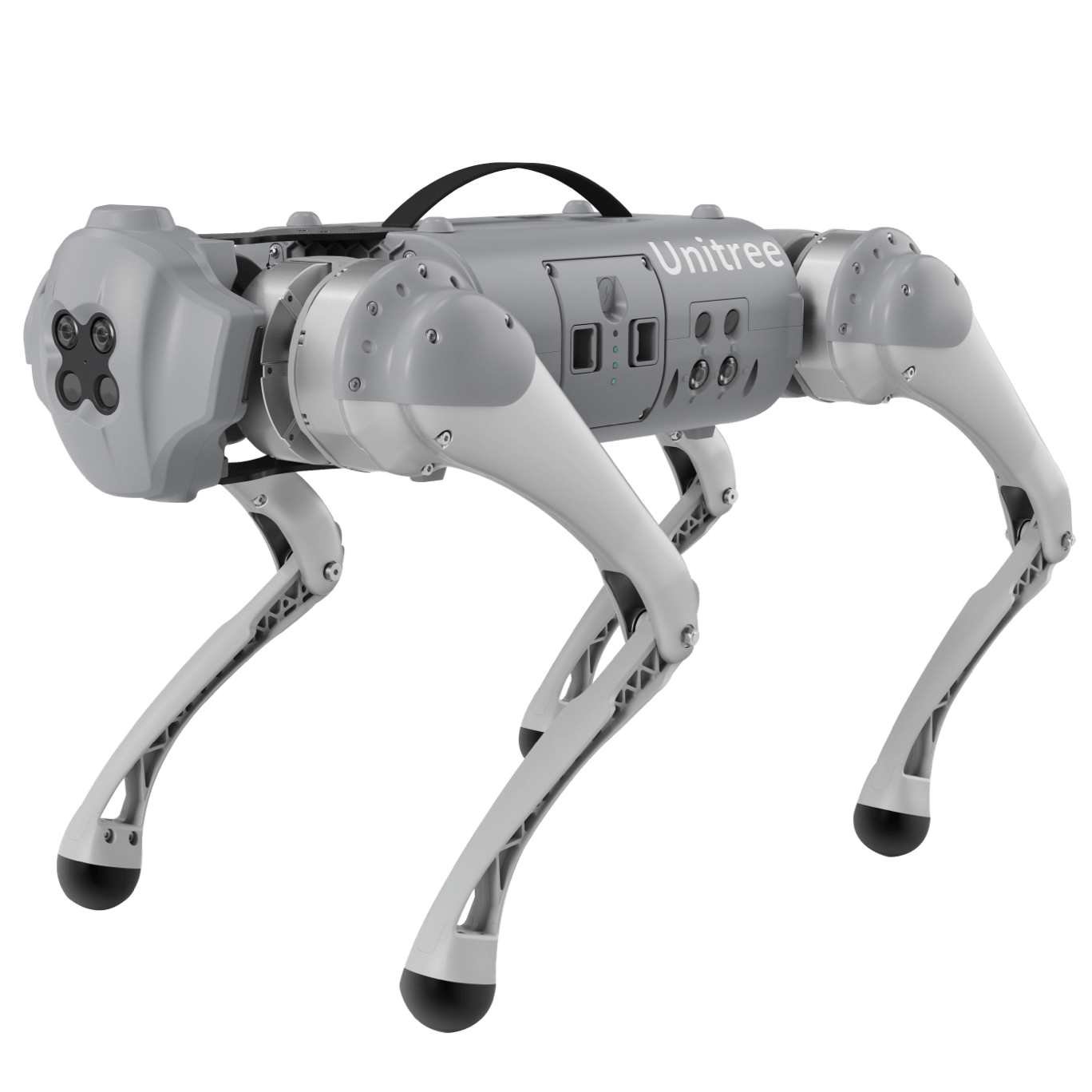}} 
    }\label{fig:Go1} %
    \subfloat[Aliengo]{{\includegraphics[width=2.5cm]{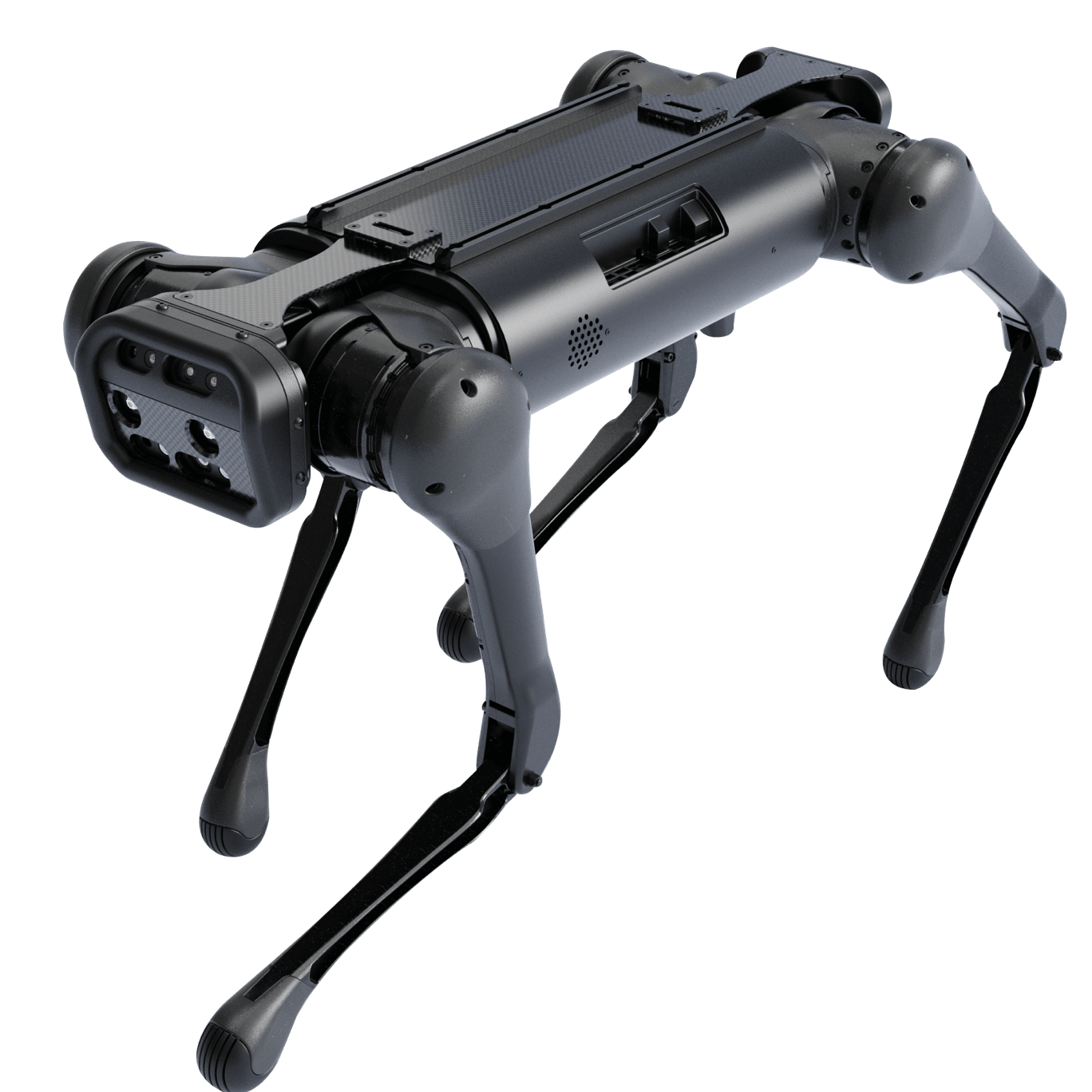}} }\label{fig:aliengo}
    \subfloat[Spot]{{\includegraphics[width=2.5cm]{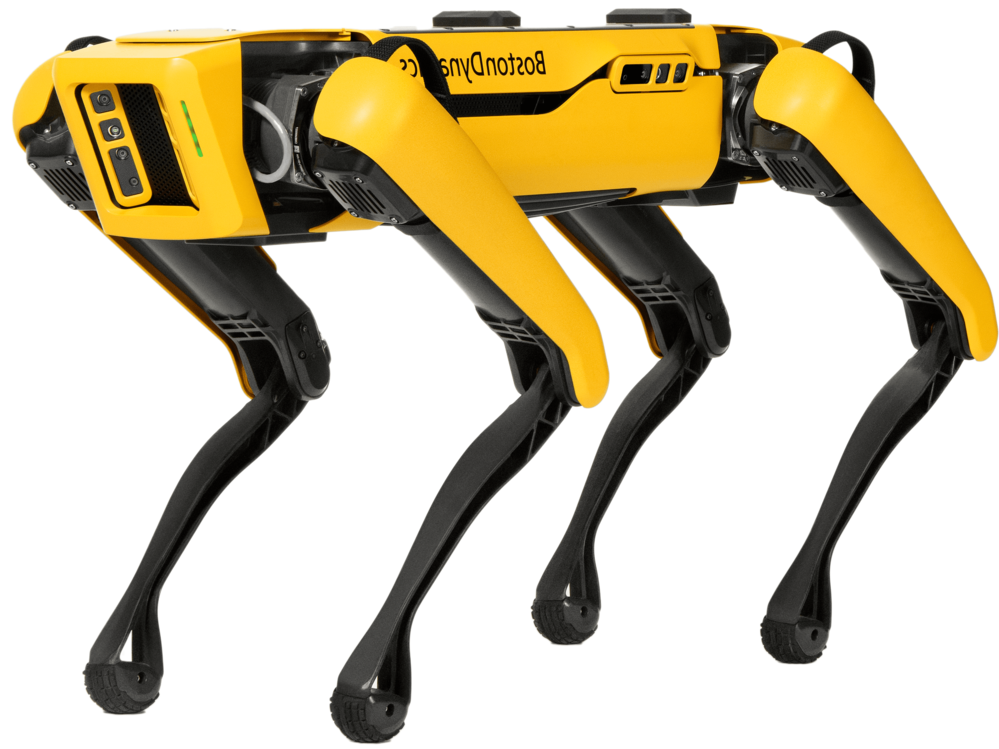}} 
    }\label{fig:spot} %
    \subfloat[B1]{{\includegraphics[width=2.5cm]{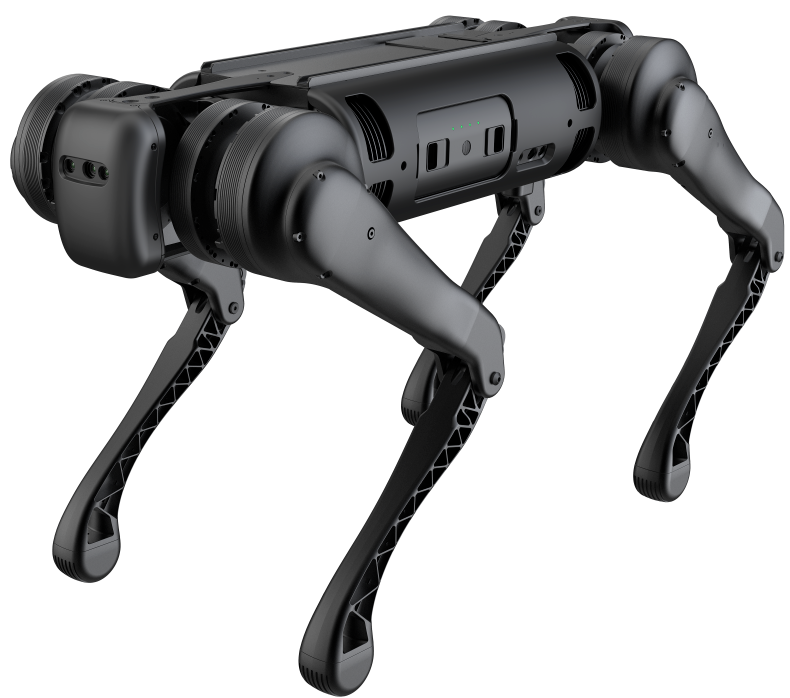}}
    }\label{fig:B1} %
    \subfloat[ANYmal]{{\includegraphics[width=2.5cm]{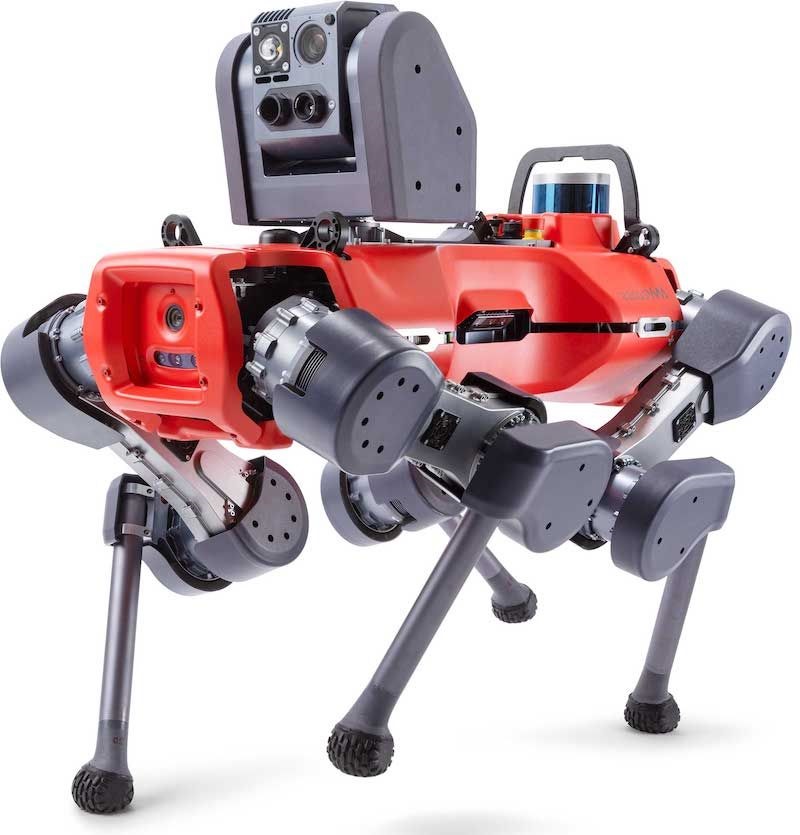}}
    }\label{fig:Anymal} %
    \caption{Currently in Europe available quadruped robots (Mai, 2023)}%
    \label{fig:available_quadruped_robots}%
\end{figure}

Figure \ref{fig:available_quadruped_robots}, tables \ref{tab:comparison1} and \ref{tab:comparison2} present the list of robots analyzed in this research together with their main properties.
The table is organized by robot weight in ascending order. 
Some indices were calculated similarly as by \citeauthor{yao2021design} (\citeyear{yao2021design}).

\begin{table}[htb]
\begin{tabular}{llccccccc}
\rowcolor[HTML]{4472C4} 
{\color[HTML]{FFFFFF} \textbf{Name}} & \multicolumn{1}{c}{\cellcolor[HTML]{4472C4}{\color[HTML]{FFFFFF} \textbf{Company}}} & {\color[HTML]{FFFFFF} \textbf{\begin{tabular}[c]{@{}c@{}}Rel. \\ Year\end{tabular}}} & {\color[HTML]{FFFFFF} \textbf{\begin{tabular}[c]{@{}c@{}}BL \\ (m)\end{tabular}}} & {\color[HTML]{FFFFFF} \textbf{\begin{tabular}[c]{@{}c@{}}H\\  (m)\end{tabular}}} & {\color[HTML]{FFFFFF} \textbf{\begin{tabular}[c]{@{}c@{}}W\\ (kg)\end{tabular}}} & {\color[HTML]{FFFFFF} \textbf{\begin{tabular}[c]{@{}c@{}}PL\\      (kg)\end{tabular}}} & {\color[HTML]{FFFFFF} \textbf{\begin{tabular}[c]{@{}c@{}}PLC\\  (\%)\end{tabular}}} & {\color[HTML]{FFFFFF} \textbf{IP}} \\
\rowcolor[HTML]{D9E1F2} 
 A1                                                                               & Unitree                                                                             & 2020                                                                                 & 0,50                                                                              & 0,40                                                                             & 12                                                                               & 7                                                                                      & 58,3                                                                                & -                              \\
Go1                                                                              & Unitree                                                                             & 2021                                                                                 & 0,65                                                                              & 0,40                                                                             & 12                                                                               & \cellcolor[HTML]{FFFFFF}5\footnote{Some specifications say that the maximum payload of the Go1 Edu is 10 kg.}                                                              & 41,7                                                                                & -                              \\
\rowcolor[HTML]{D9E1F2} 
Aliengo                                                                          & Unitree                                                                             & 2019                                                                                 & 0,65                                                                              & 0,60                                                                             & 20                                                                               & 13                                                                                     & 65,0                                                                                & -                              \\
Spot                                                                             & Boston Dyn.                                                                         & 2020                                                                                 & 1,10                                                                              & 0,61                                                                             & 32                                                                               & 14                                                                                     & 43,8                                                                                & 54                              \\
\rowcolor[HTML]{D9E1F2} 
B1                                                                               & Unitree                                                                             & 2021                                                                                 & 1,10                                                                              & 0,67                                                                             & 50                                                                               & 40                                                                                     & 80,0                                                                                & 68                              \\
ANYmal                                                                           & Anybotics                                                                           & 2019                                                                                 & 0,93                                                                              & 0,89                                                                             & 50                                                                               & 23                                                                                     & 46,0                                                                                &  67                              \\
\caption{\label{tab:comparison1} Comparison of quadruped robots. Rel. Year: Release Year; BL: body length (m); H: robot height while standing (m); W: robot weight (kg) without additional payload; PL: max play load (kg); PLC: payload capacity (\%) payload/weight; IP: Ingress protection.}
\end{tabular}
\end{table}
\begin{table}[htb]
\begin{tabular}{lcccccclcc}
\rowcolor[HTML]{4472C4} 
{\color[HTML]{FFFFFF} \textbf{Name}} & {\color[HTML]{FFFFFF} \textbf{\begin{tabular}[c]{@{}c@{}}V \\ (m/s)\end{tabular}}} & {\color[HTML]{FFFFFF} \textbf{NS}} & {\color[HTML]{FFFFFF} \textbf{NWC}} & {\color[HTML]{FFFFFF} \textbf{\begin{tabular}[c]{@{}c@{}}SS\\  (cm)\end{tabular}}} & {\color[HTML]{FFFFFF} \textbf{\begin{tabular}[c]{@{}c@{}}S \\ (deg)\end{tabular}}} & {\color[HTML]{FFFFFF} \textbf{T (h)}} & \multicolumn{1}{c}{\cellcolor[HTML]{4472C4}{\color[HTML]{FFFFFF} \textbf{M}}} & {\color[HTML]{FFFFFF} \textbf{\begin{tabular}[c]{@{}c@{}}Min P \footnote{All prices are for the german market for May 2023.}    \\  (Tsd. €)\end{tabular}}} & {\color[HTML]{FFFFFF} \textbf{\begin{tabular}[c]{@{}c@{}}Max P  \\ (Tsd. €)\end{tabular}}} \\
\rowcolor[HTML]{D9E1F2} 
A1                                  & 3,3                                                                                & 6,6                                & 385,0                               & 12                                                                                & 35                                                                                 & 1 - 2,5                               & \href{https://www.trossenrobotics.com/Shared/Unitree%20Robot/unitree-robotics-introduction-products-line_v1.2_en.pdf}{\ExternalLink}                                             & 13,5                                                                                       & -                                                                                          \\
Go1                                  & 3,7\footnote{Some specifications say that the maximum speed of the Go1 Edu is 5 m/s.}                                                                             & 5,7                                & 239,0\footnote{Assuming a payload of 10 kg and a speed of 5 m/s as for the Go1 Edu, this value increases to 646.}                              & 12                                                                                & 35                                                                                 & \cellcolor[HTML]{FFFFFF}1 - 2,5       & \href{https://www.trossenrobotics.com/Shared/Unitree%20Robot/unitree-robotics-introduction-products-line_v1.2_en.pdf}{\ExternalLink}                                             & 5,6                                                                                        & 23,1\footnote{The price of the Go1 depends on the version: Air, Pro, Edu, etc.}                                                                                      \\
\rowcolor[HTML]{D9E1F2} 
Aliengo                                  & 1,5                                                                                & 2,3                                & 150,0                               & 18                                                                                & 25                                                                                 & 2,5 - 4,5                             & \href{https://www.trossenrobotics.com/Shared/Unitree%20Robot/unitree-robotics-introduction-products-line_v1.2_en.pdf}{\ExternalLink}                                             & 44,4                                                                                       & -                                                                                          \\
Spot                                  & 1,6                                                                                & 1,5                                & 63,6                                & 22                                                                                & 30                                                                                 & 1,5                                   & \href{https://support.bostondynamics.com/s/article/Robot-specifications}{\ExternalLink}                                             & 75,0                                                                                       & -                                                                                          \\
\rowcolor[HTML]{D9E1F2} 
B1                                  & 1,8                                                                                & 1,6                                & 130,9                               & 20                                                                                & 35                                                                                 & 2 - 4                                 & \href{https://www.trossenrobotics.com/Shared/Unitree%20Robot/unitree-robotics-introduction-products-line_v1.2_en.pdf}{\ExternalLink}                                             & 70,0                                                                                       & 86,9                                                                                       \\
ANYmal                                  & 1,3                                                                                & 1,4                                & 64,3                                & 25                                                                                & 30                                                                                 & 2                                     & \href{https://www.anybotics.com/anymal-technical-specifications.pdf}{\ExternalLink}                                             & 150,0\footnote{This price of the ANYmal is not from an official seller, it might be wrong. Official prices are not disclosed to the public.}                                                                                          & -   \\
\caption{\label{tab:comparison2} Continuation of Table 1.
V: maximum speed (m/s); NS: normalized speed, maximum speed/body length; NWC: normalized work capacity, normalized speed$\times$payload capacity; SS: maximum stairways step height (cm) recommended by the manufacturer; S: slope (in degrees) it can climb on a flat surface; T: run time range (h) with one battery provided by fabricator; M: external link to more information; Min P/Max P: minimum and maximum price (€) in Germany.}  
\end{tabular}
\end{table}

Depending on diverse on-site conditions, one robot can be more suitable than another. 

\Cref{tab:comparison1} shows that it is possible to separate the listed robots into two groups: the ones with and those without water (ingress) protection.
Only the robots with ingress protection (IP) of 67 or above can be exposed to heavy rain or submerged in the water.

The maximum stairway step height is decisive when developing a completely autonomous system for construction monitoring on multiple stories.
In general, high staircase steps remain a challenge. 
While some studies have been trying to push to the maximum limits of the capabilities of different robots, like the A1 \cite{yang2023neural}, these attempts still need to improve stability. 
Assuming a standard step size of 19 cm, only the Spot, B1, and ANYmal robots would be suitable candidates.

Another essential aspect to consider is the weight of the robot when the robot needs to be repositioned manually.
Since specific robot dogs are significantly heavier, at least two people are required to carry these last three robots.

Among the compared robots, the A1 archives the highest normalized work capacity (NWC). 
However, considering the maximum specifications of the Go1 Edu, having an NWC of 646, five times more than every IP-protected robot.

The index suggests that these robots are suitable choices for quickly and effectively carrying out the scanning process.
Assuming the robot's maximum step size limitation, a possible solution is to place a robot for each level on the construction site.

As Table \ref{tab:comparison2} indicates, robot dogs' battery capacities averages are at 2.25 hours which represents a limitation for scanning large on-site environments.

    
Based on the analysis presented in this section, the Go1 Edu robot emerges as a good trade-off between the requirements of the proposed use case and price.

\section{Data acquisition process}

Among the different current available legged robots presented in this paper, we conducted our case study with the \textit{Go1 Edu} from the Unitree company.

As our main purpose is to leverage the capabilities of the robot for autonomous navigation and mapping, we developed a mapping system that can be used independently of the robot. 
In this way, the system can also be operated as a handheld or over any other robot.


Since having different sensor modalities contributes to achieving accurate pose estimation and, therefore a more accurate map acquisition,  we developed a system that integrates light detection and ranging (LiDAR), Camera, and inertial measurement unit (IMU) sensors. 
For our robot-independent system, we chose to use the ASRock 4x4 BOX-5800U mini personal computer equipped with 32 GB of RAM, together with two batteries XTPower XT-27000 DC-PA. 

The most important criteria for selecting the PC were a high-performance CPU and low power consumption. 

\begin{figure}[hbt]
    \caption{Developed mobile mapping system. (a) The system is placed over the Go1 robot, with the help of the custom-designed mounting system which allows the montage on any robot with a flat surface and also allows the usage of the system as a handheld; (b) corresponding connection and data transfer diagram of the mapping system}%
    \label{fig:go1}%
    \includegraphics[width=\textwidth]{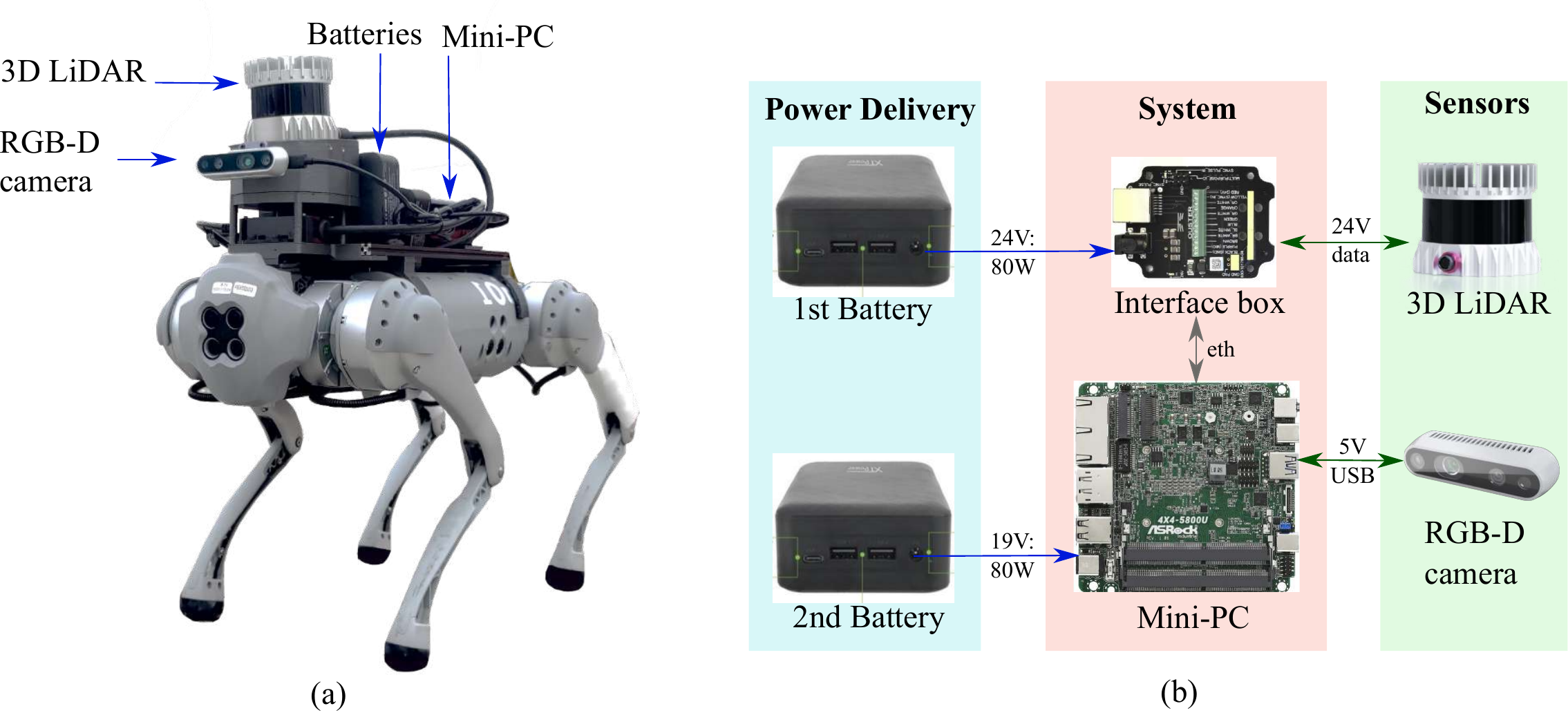}%
\end{figure}

Figure \ref{fig:go1} shows the Go1 robot dog equipped with the developed mapping system and the schematic connection between the different components.

The mounting system that allows the attachments of the sensor over the robot was designed considering mainly two requirements: 
First, maintain a low center of gravity, and second, use as less material as possible to keep the system light (< 5 kg) and portable by the robot.

The system comprises twelve 3D-printed custom-design parts, two metal maker beams, and several screws and inserts.
The fact that the system has several parts makes it modular and allows easy maintenance, for example, if one part is broken can be easily substituted.

Moreover, the shape of the different parts considers the system's possible usage above other robots (with a flat surface) or as a handheld system. 
To use it as a handheld system, one can easily separate the LiDAR and camera and add them to a handle, and the mini-PC and battery could be placed on a backpack.

\begin{figure}[!htb]%
    \centering
    \subfloat[Robots view using object detection to understand the environment.]{{\includegraphics[width=6cm]{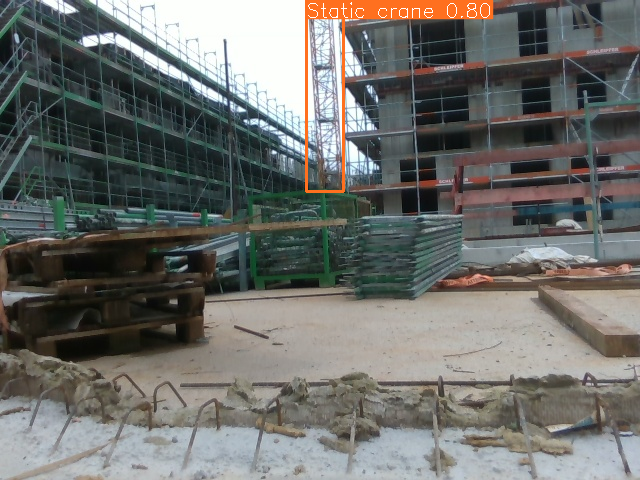}}\label{fig:robot_view} }
    \subfloat[Point cloud reconstructed from the measurements of the sensors over the robot with a SLAM system.]{{\includegraphics[width=6cm]{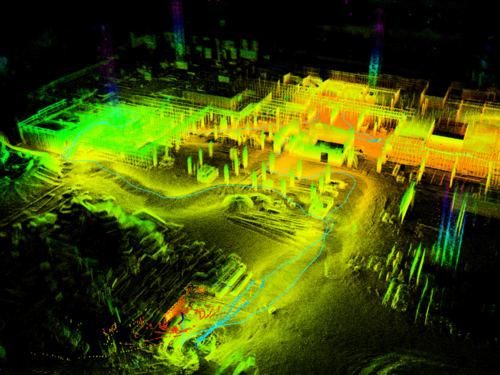}} \label{fig:point_cloud}} \\%
    \subfloat[Bird-view of the robot and construction worker detected by the UAV's camera.]{{\includegraphics[width=12cm]{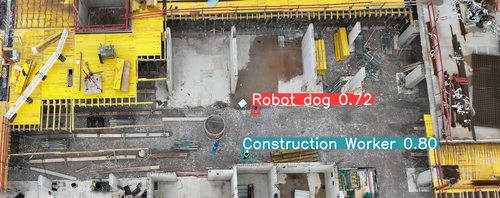}}\label{fig:drone} } %
    \caption{Real-world on-site experiments with the robot dog.}%
    \label{fig:robot_data_acquiisition}%
\end{figure}

For the data acquisition process, we follow the next steps.
(1) all the sensors and mini-PC are connected to the power delivery; 
(2) a remote connection with the mini-PC through a remote desktop application is established; 
(3) the software development kit (SDK) of each sensor is launched using nodes of the robot operating systems (ROS);
(4) a synchronization process between the images and the LiDAR data allows the recording of LiDAR scans and camera images with the same time stamps at 10 Hz of frequency.
(5) a SLAM system, specifically an enhanced version of FAST-LiDAR inertial odometry (LIO) \cite{FAST-LIO2_Xu.2022} with loop closure capabilities is leveraged to create 3D maps of the environment in real-time. 
The data acquisition was conducted on a building construction site within the Munich area, covering a total area of 12.500 square meters. Figure \ref{fig:point_cloud} illustrates the resulting 3D point cloud.

The acquired data consist of sequential images, LiDAR scans, and IMU measurements.

While total autonomy is still in development, a BIM model can be leveraged to navigate the robot autonomously in controlled environments using the ROS navigation stack as explained in \cite{vega:2022:2DLidarLocalization}.

\subsection{Analysis of acquired data}

Once the 3D data is acquired, it can subsequently be aligned, corrected, and analyzed with the support of a BIM model, as explained in \cite{vega:2023:BIM_SLAM}. 
After this process, all the data, even if they were acquired at different time stamps, should be in the same coordinate system. This means, that the camera images as well as the LiDAR scans should have known poses aligned with the BIM model.

Further, automatic semantic enrichment of 3D point cloud is possible with the method proposed here \cite{Vega:2022:ObjectDetection}. 
This method would allow the detection of cranes, scaffolding, and formwork, which are elements that are very often present on construction sites.

On the other hand, the gained imaged data can be processed further by object detection pipelines and linked to other components in the construction environment, as explained more in detail here \cite{Pfitner:2023:ObjectDetection}.
An example image from the robot's view with a detected crane is shown in figure \ref{fig:robot_view}.
In this case, the robot dog extends state-of-the-art monitoring methods for indoor areas, which are not covered by crane cameras or UAVs \cite{Collins:2022:PC}.
Considering the unsolved imperfections of current digital twins aiming to cover the entire construction site \cite{Schlenger:2023:DT}, the robot dog presents a promising data acquisition extension.

\section{Discussion}
Bringing the Go1 robot dog to construction sites poses several challenges. While the reliability for manual operation is generally sufficient, issues arise when attempting autonomous navigation due to software update problems and mapping inaccuracies in complex environments with moving objects. Certain areas on construction sites become unreachable for the robot due to high stairway steps or slopes, requiring manual carrying. 

Additionally, the Go1 lacks adequate resistance to harsh weather conditions, hindering further experiments in adverse environments. The limited battery capacity, lasting only around 30 minutes of exhaustive use, necessitates manual charging efforts. To make semi-autonomous deployment feasible, certain requirements must be met, including the availability of comprehensive and reliable 3D BIM models or maps, continuous updating of major construction site changes on the map, handling deployment level-wise to minimize barriers, implementing precise and fast path planning algorithms for autonomous navigation, and significantly improving battery capacity through manufacturer enhancements and employing an autonomous charging method.

\section{Conclusion}
This study explored the practicality of employing the currently available robot dogs in construction sites, with a specific emphasis on their effectiveness in enabling data acquisition.

In addition, we conducted a real-world experiment on a large-scale construction site using a quadruped robot equipped with a self-developed mapping system.

We can conclude that robot dogs are suitable for scanning construction sites on a frequent basis, specifically indoor environments. 
Robot dogs can extend current monitoring solutions by providing valuable semantic and geometric data, facilitating the creation of a digital twin.
However, the following limitations must be addressed prior to a feasible deployment: their limited battery capacity, their lack of adaptability to dynamic and harsh environments, and their prototypical condition. 
We argue that the deployment of multiple robot dogs can overcome some of the current limitations.
In summary, we believe that robot dogs are a valuable tool for monitoring complex construction environments in the future, specifically when technical improvements diminish their limitations. 

\section*{Acknowledgements}
We thankfully acknowledge Innovation Management Bau GmbH for their financial support and for providing us access to multiple construction sites to collect valuable data.

The presented research was conducted in the frame of the project ``Intelligent Toolkit for Reconnaissance and assessmEnt in Perilous Incidents'' (INTREPID) funded by the EU's research and innovation funding programme Horizon 2020 under Grant agreement ID: 883345.

\printbibliography

\label{sec:ref}

\end{document}